\documentclass[letterpaper, 10 pt, conference]{ieeeconf}
\IEEEoverridecommandlockouts

\usepackage{amsmath}
\usepackage{cite}
\usepackage{amssymb}
\usepackage[ruled,linesnumbered,noline]{algorithm2e}

\usepackage{float}
\usepackage{textcomp}
\usepackage{bm}

\newtheorem{thm}{Theorem}

\newtheorem{definition}{Definition}
\newtheorem{problem}{Problem}
\newtheorem{assum}{Assumption}
\newtheorem{subproblem}{Subproblem}[problem]

\usepackage{hyperref}

\DeclareFontFamily{U}{matha}{\hyphenchar\font45}
\DeclareFontShape{U}{matha}{m}{n}{
      <5> <6> <7> <8> <9> <10> gen * matha
      <10.95> matha10 <12> <14.4> <17.28> <20.74> <24.88> matha12
      }{}
\DeclareSymbolFont{matha}{U}{matha}{m}{n}

\DeclareMathSymbol{\Lt}{3}{matha}{"CE}
\DeclareMathSymbol{\Gt}{3}{matha}{"CF}

\usepackage{graphicx}
\usepackage{tikz}
\usepackage{float}
\usetikzlibrary{arrows}
\usetikzlibrary{arrows.meta}
\usetikzlibrary{arrows,positioning} 
\usetikzlibrary{shapes}
\usetikzlibrary{calc}
\usetikzlibrary{fadings}
\usetikzlibrary{decorations.pathreplacing}
\usetikzlibrary{decorations.text}
\usepackage{makecell}
\usepackage[T1]{fontenc}

\usepackage{amsfonts}

\def\eventually{\Diamond}
\def\until{\mathcal{U}}

\usepackage{scalerel}
\def\msquare{\mathord{\scalerel*{\Box}{gt}}}
\def\mdiamond{\mathord{\scalerel*{\Diamond}{gt}}}
\makeatletter \providecommand\dotdiamond{\mathpalette\@barred\mdiamond} \def\@barred#1#2{\ooalign{\hfil$#1\cdot$\hfil\cr\hfil$#1#2$\hfil\cr}}  \makeatother
\makeatletter \providecommand\dotbox{\mathpalette\@burrow\msquare} \def\@burrow#1#2{\ooalign{\hfil$#1\cdot$\hfil\cr\hfil$#1#2$\hfil\cr}}  \makeatother

\tikzfading[name=fade right,
left color=transparent!0, right color=transparent!100]
\tikzfading[name=fade out, 
    inner color=transparent!0,
    outer color=transparent!100]
\definecolor{dcompb}{RGB}{157,35,0} 
\definecolor{mblue}{RGB}{0.176, 0.380, 0.659}
\definecolor{dgreen}{RGB}{15,111,3}
\definecolor{dred}{RGB}{.65,.176,0}
\definecolor{dblue}{RGB}{.098,.243,.424}
\usepackage[top=54pt, left=54pt, right=54pt, bottom=54pt]{geometry}
\begin{document}

\title{\textbf{\huge A Unified Approach to Multi-task Legged Navigation: Temporal Logic Meets Reinforcement Learning}\vspace*{3pt}}

\author{Jesse~Jiang, Samuel~Coogan, and Ye~Zhao\thanks{This work was supported in part by the National Science Foundation under grant \#1924978 and by the National Science Foundation Graduate Research Fellowship under grant \#DGE-2039655.}
\thanks{Jesse Jiang and Samuel Coogan are with the School of Electrical and Computer Engineering, Georgia Institute of Technology, Atlanta, GA 30332 USA (e-mail: jjiang@gatech.edu, sam.coogan@gatech.edu). S. Coogan is also with the School of Civil and Environmental Engineering. }
\thanks{Ye Zhao is with the School of Mechanical Engineering, Georgia Institute of Technology, Atlanta, GA 30332 USA (e-mail: ye.zhao@me.gatech.edu).}}

\maketitle

\thispagestyle{empty}
\begin{abstract}
This study examines the problem of hopping robot navigation planning to achieve simultaneous goal-directed and environment exploration tasks. We consider a scenario in which the robot has mandatory goal-directed tasks defined using Linear Temporal Logic (LTL) specifications as well as optional exploration tasks represented using a reward function. Additionally, there exists uncertainty in the robot dynamics which results in motion perturbation. We first propose an abstraction of 3D hopping robot dynamics which enables high-level planning and a neural-network-based optimization for low-level control. We then introduce a Multi-task Product IMDP (MT-PIMDP) model of the system and tasks. We propose a unified control policy synthesis algorithm which enables both task-directed goal-reaching behaviors as well as task-agnostic exploration to learn perturbations and reward. We provide a formal proof of the trade-off induced by prioritizing either LTL or RL actions. We demonstrate our methods with simulation case studies in a 2D world navigation environment.
\end{abstract}

\section{Introduction}
The problem of complex navigation planning for robotic systems has increasingly gained relevance due to societal interest in deploying robots in real-world environments. In these scenarios, robots must execute task-directed control policies to achieve desired behaviors while also performing task-agnostic exploration to adapt to unknown environments. Furthermore, for deployment in real-world environments it is paramount to design a seamlessly integrated navigation planning framework that can handle both dynamics and environmental complexities. 

This study draws inspiration from both formal methods and reinforcement learning (RL) approaches to the navigation planning problem. Formal methods provide guarantees on satisfiability of tasks while RL algorithms provide a notion of optimal satisfaction of objectives. The goal of this study is to create a unified planning approach combining goal-reaching and exploration policies to satisfy multi-task objectives.

\begin{figure}[t]
    \centering
    \includegraphics[width=0.475\textwidth]{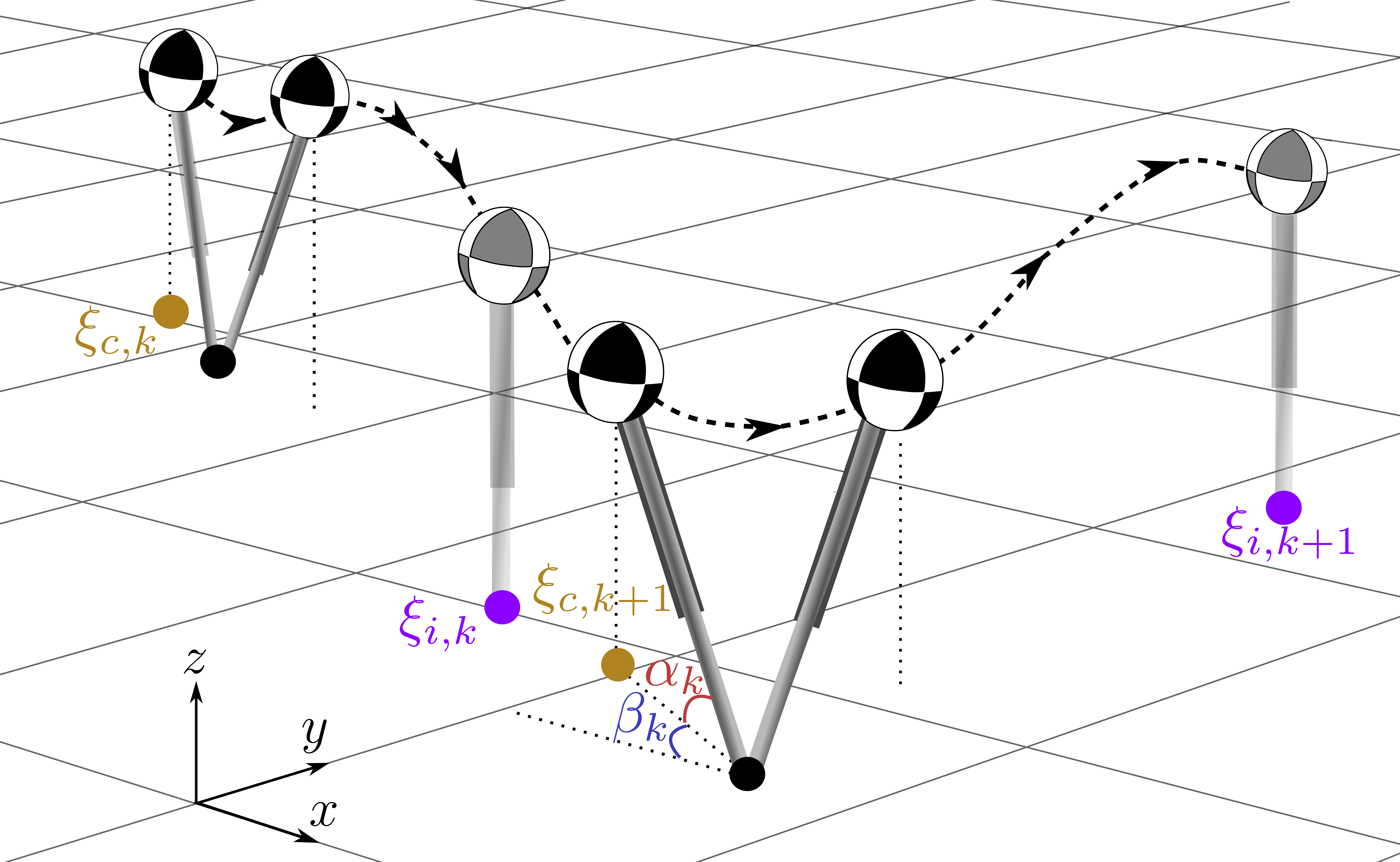}
    \caption{Illustration of the dynamic motion for hopping robots considered in this work. At the initial contact state $\xi_{c,k}$, the trajectory of the next hop can be determined via forward simulation of the dynamics until the interstitial state $\xi_{i,k}$. The objective of the learned controller is to achieve the desired next interstitial state $\xi_{i,k+1}$ by controlling the leg angles $\alpha_k,\beta_k$ at the next contact state $\xi_{c,k+1}$. }
    \label{fig:Two Hop Robot}
    \vspace{-0.2in}
\end{figure}
\subsection{Related Work}
In both the RL and formal methods literature, avenues for combining hard (mandatory) and soft (optional) tasks have been explored. One RL approach uses constrained MDPs \cite{geibel2006,wachi2020safe,gattami21a} to incorporate hard and soft constraints. This approach maximizes the reward from soft constraints as in standard RL algorithms. Additionally, a constraint is added in the optimization problem to enforce that the reward earned from a reward function encoding the hard constraints exceeds some threshold value. 
Another well-received RL approach is the "shielding" method \cite{alshiekh2018safe,De_Giacomo_Iocchi_Favorito_Patrizi_2021}. 
In this formulation, the agent runs a traditional RL exploration/exploitation algorithm, but the control action taken is modified by a higher-level shield to prevent the agent from entering states violating a syntactically co-safe LTL (scLTL) specification.

Temporal logic approaches to the combined hard and soft constraint problem focus on performing optimization of soft rewards over a set of computed control policies which satisfy the hard temporal logic constraints. The work of \cite{ding2011mdp} studies the problem of minimizing the cost of an infinite cycle control policy which satisfies given LTL specifications. However, the transient cost of reaching an accepting state is not accounted for. In \cite{wen2015}, the first approach is proposed for simultaneously ensuring satisfaction of LTL specifications while optimizing a reward function using RL. The method approximates a maximal set of control policies for a deterministic two-player game which is guaranteed to satisfy the LTL specifications. Then, maximin Q-learning optimizes over this set of policies. However, the approach is only valid for nonprobabilistic systems and guarantees of optimality are shown for a limited fragment of LTL.

From a robot locomotion perspective, temporal-logic-based planning methods for legged robots haveß been underexplored, with a few exceptions of bipedal and quadrupedal locomotion \cite{shamsah2023integrated,gu2023walkingbylogic}. The work of \cite{sakamoto2022hopping} proposes an MDP-based motion planner for a hopping robot in uncertain environments, but does not integrate tightly with the underlying robot dynamics. The works in \cite{wu2013,He2008modeling, arslan2011reactive, piovan2015reachability, yim2018precision} develop motion planners and controllers for hopping robots compatible with the underlying low-level dynamics, but their control objectives are associated with stability rather than high-level navigation tasks.

\subsection{Contributions}
In this study, we design a unified planning method for a hopping robotic system which ensures satisfaction of mandatory tasks specified by Linear Temporal Logic (LTL) and maximizes the reward gained from optional tasks. We propose a Multi-task Product IMDP (MT-PIMDP) model which abstracts the system dynamics and both classes of tasks. We also develop a neural-network-based optimization to execute high-level plans on the full-order system.

Our contributions are as follows:
\begin{itemize}
    \item We propose the first probabilistic planning approach to satisfy goal-reaching tasks specified using LTL while simultaneously maximizing RL rewards for exploration. 
    \item We develop a novel IMDP abstraction and neural-network-based optimization for 3D hopping robot dynamics. This model enables high-level navigation planning for multiple tasks under system uncertainty while maintaining low-level kinodynamic feasibility.
    \item We provide a formal proof on the trade-off induced by our planning policy between the number of steps required to satisfy the LTL tasks and the amount of reward gained from optional tasks.
\end{itemize}
\section{Preliminaries}\label{section:Setup}
\subsection{SLIP Hopping Model}\label{subsection: SLIP Model}
We consider a 3D hopping robot modeled as a point mass $m$ with a spring leg, similar to \cite{wu2013}. During the flight phase, the point mass is subject to gravity only and follows a ballistic trajectory as
\begin{equation}\label{eq: Flight Phase}
    m\begin{bmatrix}
        \ddot{x} & \ddot{y} & \ddot{z}
    \end{bmatrix}^\top = \begin{bmatrix}
        0 & 0 & -mg
    \end{bmatrix}^\top
\end{equation}
During the stance phase, we assume that the robot rebounds elastically according to the following equation
\begin{equation}\label{eq: Impact Phase}
    m\begin{bmatrix}
        \ddot{x} \\ \ddot{y} \\ \ddot{z}
    \end{bmatrix} = (\frac{kl_0}{|\bm{l}|}-1)\begin{bmatrix}
        x-x_f \\ y-y_f \\ z
    \end{bmatrix}+
    \begin{bmatrix}
        0 \\ 0 \\ -mg
    \end{bmatrix}+\bm{f}(\begin{bmatrix}
        x-x_f \\ y-y_f \\ z
    \end{bmatrix})
    +\bm{\nu}
\end{equation}
where $k$ is the spring constant, $l_0$ is the rest length of the leg, $\begin{bmatrix}
    x_f & y_f & z
\end{bmatrix}$
is the 3D location of the foot point, the leg length vector $\bm{l} = $ $\begin{bmatrix}
        x-x_f & y-y_f & z
    \end{bmatrix}^\top$, $|\bm{l}|$ is the leg length, $\bm{f}$ is an unknown state-dependent perturbation caused by, \textit{e.g.}, terrain uncertainty, and $\bm{\nu} \in \mathbb{R}^3$ is additive noise.

We now define the \textit{interstitial state} of each hop, which represents the moment that the robot would contact the ground if its leg was perpendicular to the ground. 
\begin{definition}[Interstitial State]
    The interstitial state of a hop is the $(x,y)$ position of the robot center of mass (CoM) at the moment when the conditions $y=l_0$ and $\dot{y}<0$ are satisfied (the dashed lines in Figure \ref{fig:Two Hop Robot}).
\end{definition}

In this work, we will define the $k^{\rm th}$ hop of the robot as beginning at the state $\xi_{c,k}$ when the robot's leg contacts the ground and ending at the next contact $\xi_{c,k+1}$. We denote the interstitial state which occurs between the two contacts as $\xi_{i,k}$. The velocity of the robot at the $k^{\rm th}$ interstitial state is $\bm{v}_{i,k}$. 
Figure \ref{fig:Two Hop Robot} illustrates the robot states. 

Note that from the contact state $\xi_{c,k}$, the trajectory of the robot can be simulated until the interstitial state $\xi_{i,k}$ regardless of the leg angles $\alpha_k,\beta_k$ of the robot at the next contact state $\xi_{c,k+1}$. This can be seen from the system dynamics \eqref{eq: Flight Phase}, \eqref{eq: Impact Phase}, where the only control input is the foot contact location.
We propose the use of interstitial states as the discrete states on which our planner operates. One benefit is that the interstitial state $\xi_{i,k}$ can be calculated once the robot reaches the previous contact state $\xi_{c,k}$, allowing more time for the high-level navigation planner to compute a desired high-level waypoint $\xi_{i,k+1}$ and corresponding leg angles $\alpha_k,\beta_k$. Additionally, the interstitial state $\xi_{i,k}$ is invariant with respect to the next low-level control input $\alpha_k,\beta_k$ (whereas the contact state $\xi_{c,k+1}$ is not), decoupling the problems of high-level planning and low-level control.

For hopping step $k$ beginning at a contact state $\xi_{c,k}$, we simulate the trajectory of the robot until the following interstitial state $\xi_{i,k}$. Then, a high-level planner selects a targeted interstitial state $\xi_{i,k+1}$, and a low-level planner computes the corresponding contact leg angles $\alpha_k,\beta_k$. Once the robot makes contact at state $\xi_{c,k+1}$, the cycle repeats.

\subsection{Temporal Logic Planning}
The mandatory tasks are represented using the class of Linear Temporal Logic (LTL) specifications \cite[Def. 2.1]{belta_formal_2017},

which can be checked using deterministic Rabin automata (DRA) \cite[Def. 2.7]{belta_formal_2017}.
\begin{definition}[Deterministic Rabin Automaton]
A DRA is a tuple $\mathcal{R}=(S,s_0,O,\delta,F)$, where
\begin{itemize}
\item $S$ is a finite set of states,
\item $s_0\subset S$ is a singleton initial state,
\item $O$ is the input alphabet, which corresponds to observations from the LTL formula,
\item $\delta:S\times O\rightarrow 2^S$ is a transition map which is either $\emptyset$ or a singleton for all $s\in S$ and $o\in O$, and
\item $F=\{(G_1,B_1),\cdots,(G_n,B_n)\}$, where $G_i,B_i\subseteq S,i=1,2,\cdots,n$ is the acceptance condition.
\end{itemize}

\end{definition}

We next introduce Interval Markov Decision Processes:
\begin{definition}[Interval Markov Decision Process]
An \emph{Interval Markov Decision Process (IMDP)} is a tuple $\mathcal{I} = (Q, A, \check{T}, \hat{T},Q_0, O, L)$ where:
\begin{itemize}
    \item $Q$ is a finite set of states,
    \item $A$ is a finite set of actions,
    \item $\check{T},\hat{T}: Q \times A \times Q' \xrightarrow{} [0,1]$ are lower and upper bounds, respectively, on the transition probability from state $q \in Q$ to state $q' \in Q$ under action $\alpha \in A$,
    
    \item $Q_0 \subseteq Q$ is a set of initial states,
    \item $O$ is a finite set of atomic propositions or observations,
    \item $L: Q \xrightarrow{} O$ is a labeling function.
\end{itemize}

\end{definition}

To learn system uncertainties, we will utilize Gaussian process (GP) regression \cite{Williams96gaussianprocesses}.

In practice, we use a sparse Gaussian process regression approximation \cite{leibfried2021tutorial} to reduce computational complexity.

Verification and synthesis problems for IMDP systems evaluated against LTL specifications are often solved using automata-based synthesis methods on a product IMDP:
\begin{definition}[PIMDP]\label{def:PIMDP}
Let $\mathcal{I} = (Q, A, \check{T}, \hat{T},Q_0, O, L)$ be a IMDP and $\mathcal{A}=(S,s_0,O,\delta,F)$ be an DRA. The product IMDP (PIMDP) is defined as a tuple $\mathcal{P} = \mathcal{I}\otimes\mathcal{A} =$\\ $(Q\times S, A, \check{T}', \hat{T}', Q\times s_0, F')$, where
\begin{itemize}
    \item $\check{T}': (q,s) \times A \times (q',s') :=
        \check{T}(q,\alpha,q') $ if $ s' \in \delta (s, L(q))$ and
        $0$ otherwise,
    \item $\hat{T}': (q,s) \times A \times (q',s') :=
        \hat{T}(q,\alpha,q') $ if $ s' \in \delta (s, L(q))$ and
        $0$ otherwise,
    \item $(q_0,\delta(s_0,L(q_0))) \in (Q\times S)$ is a set of initial states of $\mathcal{I}\otimes\mathcal{A}$, and
    \item $F'=Q\times F=\{Q\times (G_1,B_1),\cdots,Q\times (G_n,B_n)\}$, where $G_i,B_i\subseteq S$ is the $i$th acceptance condition.
\end{itemize}
\end{definition}

\section{Problem Statement}
Our objective is to design a unified control policy synthesis algorithm which allows a hopping robot to learn system uncertainties to satisfy mandatory goal-reaching tasks with sufficient probability while simultaneously maximizing the reward of optional exploration tasks.
\begin{problem}\label{problem:Problem1}
Design an algorithm to learn the unknown dynamics of the system modeled using Equations \eqref{eq: Flight Phase} and \eqref{eq: Impact Phase} to satisfy mandatory tasks $\phi$ specified using LTL while simultaneously maximizing the reward $W$ of optional tasks.
\end{problem}

To solve Problem 1, we need to address three subproblems. We first develop an IMDP abstraction of the hopping robot dynamics which maps between the low-level 3D hopping dynamics and high-level waypoint planning.
\begin{subproblem}\label{subproblem:Abstract}
    Develop an abstraction of the hopping robot dynamics in Equations \eqref{eq: Flight Phase} and \eqref{eq: Impact Phase} which plans with respect to high-level waypoints and maps to feasible actions on the low-level dynamical system of the hopping robot.
\end{subproblem}

We then create a product IMDP framework which combines the IMDP with the LTL and reward tasks.
\begin{subproblem}\label{subproblem:MT-PIMDP}
    Design a product IMDP model which contains the IMDP abstraction of hopping robot dynamics, the LTL specifications, and the reward function task.
\end{subproblem}

We finally develop a synthesis algorithm which produces a control policy solving both the LTL and the RL tasks.
\begin{subproblem}\label{subproblem:Planning}
Synthesize a control policy for the robot which satisfies mandatory tasks $\phi$ while optimizing the known or unknown reward function $W$ of the optional tasks.
\end{subproblem}

\section{IMDP Abstraction of Hopping Robot}
In this section, we elucidate a novel abstraction of hopping robot dynamics. A key component of our abstraction is the development of a neural-network-based optimization which enables the robot to track high-level waypoints while providing a notion of low-level dynamical feasibility. Figure \ref{fig:Overall Block Diagram} depicts the structure of our framework.
\subsection{State Abstraction}
We first introduce our IMDP abstraction of the hopping robot dynamics in Equations \eqref{eq: Flight Phase} and \eqref{eq: Impact Phase}. To begin with, we partition the state space of the robot along the $x-y$ plane:
\begin{assum}
\label{assum:partitions}
The state space is bounded and is partitioned into hyper-rectangular regions $\{\Psi_q\}_{q\in Q}$ defined as
\begin{equation}
\label{eq:part}
    \Psi_q=\{\psi \mid \ a_{x,q}\leq x_\psi \leq b_{x,q},a_{y,q}\leq y_\psi \leq b_{y,q}\}\subset X,
\end{equation}
where the inequality is taken elementwise for lower and upper bounds $a_{x,q},b_{x,q},a_{y,q},b_{y,q}\in\mathbb{R}$ and $Q$ is a finite index set of the regions. Each region has a geometric center 
$c_{q}={\begin{bmatrix}
    a_{x,q}+b_{x,q}&a_{y,q}+b_{y,q}
\end{bmatrix}^\top}/{2}$. Additionally, the system possesses a labeling function $L$ which maps hyper-rectangular regions to observations $O$.
\end{assum}

We use the hyper-rectangular partitions of the state space as IMDP abstraction states. At each abstraction state, we assume that the hopping robot has the control authority of hopping up, down, left, or right in the global frame either one or two states across. Formally, we have the action set
\begin{equation}\label{eq:action}
    A=\{N_1,N_2,E_1,E_2,S_1,S_2,W_1,W_2\}
\end{equation}
where $A_j$ is a hop in the $A$ direction of $j$ states in length. This assumption is possibly violated due to the robot's dynamical constraints. In this case, we will design a backup controller in Section \ref{subsection: Low Level} to take additional hopping steps such that the robot can recover and execute the original desired action.

We can now define the IMDP abstraction of the hopping robot dynamics in Section \ref{subsection: SLIP Model}:
\begin{definition}[IMDP Abstraction of Hopping Dynamics]
    An IMDP abstraction of the hopping robot dynamics in Section \ref{subsection: SLIP Model} is an IMDP satisfying the following conditions:
    \begin{itemize}
        \item The set of states $Q$ correspond to the hyper-rectangular regions $\Psi_q$ of the robot interstitial states,
        \item The actions $A$ target the center of a hyper-rectangular region $\Psi_q$.
    \end{itemize}
\end{definition}

The IMDP abstraction discretizes the 3D robot trajectory into a hop-by-hop procedure which targets interstitial states $\xi_i$ corresponding to the centers of hyper-rectangular regions $\Xi_q$. 
At step $k$ of a hopping trajectory, the IMDP action generates a desired position for the final interstitial state $\xi_{i,k+1}$, and the desired hop is the 3D motion between $\xi_{i,k}$ and $\xi_{i,k+1}$ as illustrated in Figure \ref{fig:Two Hop Robot}.
\subsection{Learning-based Controller}\label{subsection: Learning Controller}
We now detail the controller which maps a target interstitial state to the corresponding leg angle at the next impact state, creating several challenges. 
\begin{figure}[t]
    \centering
    \includegraphics[width=0.475\textwidth]{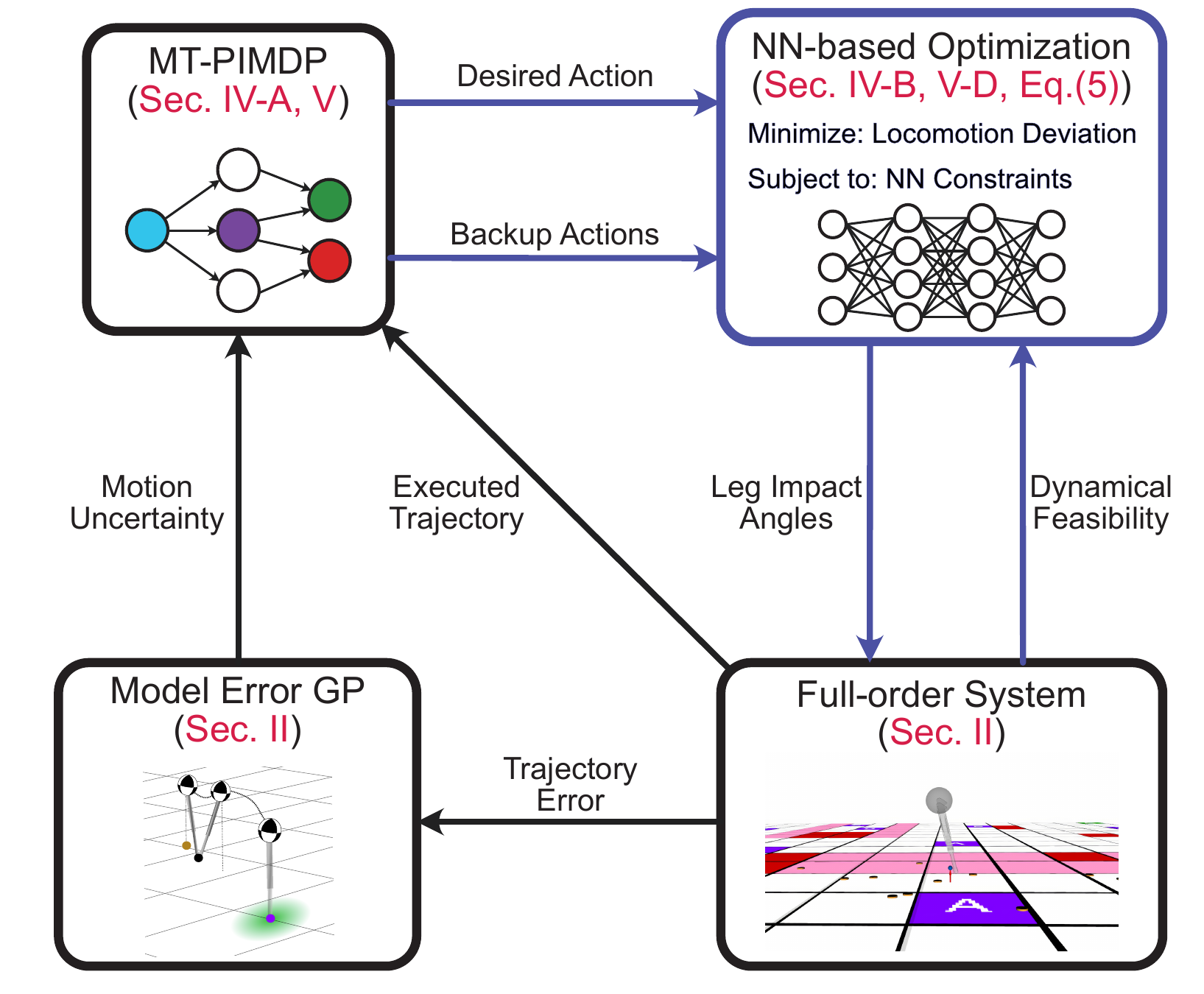}
    \caption{Overall block diagram of the framework. The NN-based Optimization block and its connections (in blue) are expanded in Figure \ref{fig:Hopping Robot Switching}.}
    \label{fig:Overall Block Diagram}
    \vspace{-0.2in}
\end{figure}

First, the timing of the impact phase is dependent on the leg angle selection, so that the control input is coupled with the phase transitions. Additionally, the controller is non-deterministic: there could exist multiple leg angles which achieve the same hop displacement, analogous to the problem of throwing a ball on a ballistic trajectory \cite{Brown_2013}.

To address these challenges, we propose a neural network controller to learn the controller. The multi-layer perceptron (MLP) architecture used in this work learns the coupling between the control input and the impact phase timing, solving the first challenge, but it does not learn non-deterministic output features effectively. To address this second challenge, we use the MLP to learn the reverse mapping from the leg angle to the hop displacement. Formally, the inputs to the neural network are the CoM velocity at the initial interstitial state $\xi_{i,k}$ and the leg impact angles $\alpha,\beta$, and the outputs are the hop displacement $\xi_{i,k+1}-\xi_{i,k}$ and CoM velocity at the final interstitial state $\xi_{i,k+1}$. The MLP is trained offline using data collected by forward simulating the hopping robot dynamics from a range of initial interstitial state velocities ($[1,8]$ m/s) and leg angles ($\alpha\in[\frac{\pi}{4},\frac{\pi}{2}],\beta\in[-\frac{\pi}{2},\frac{\pi}{2}]$)\footnote{The velocity of the hopping robot varies greatly as a result of the need to change directions quickly, so a large range of input data parameters is needed to account for the possible configurations observed at runtime.}.

At runtime, we need a controller which performs the reverse mapping from the hop displacement to the corresponding leg angle. We obtain such a controller by optimizing over the input space of the neural network the problem 
\begin{align}\label{eq: NN Optimization}
    \nonumber\arg\min_{\alpha_k,\beta_k} &\ c_1||\xi_{i,k+1}-\xi_{\text{des},k+1}||_2 + c_2 ||\bm{v}_{i,k+1}-\bm{v}_{\rm des}||_2 \\
    &+ c_3{\rm Bound}(\alpha_k,\beta_k) \\
    \nonumber\text{s.t.} &\ (\xi_{i,k+1}-\xi_{i,k}, \bm{v}_{i,k+1})= NN(\bm{v}_{i,k},\alpha_k,\beta_k)
\end{align}
where $\xi_{\text{des},k+1}$ is the target interstitial state; constants $c_1,c_2$ and $c_3$ are weighting coefficients, $c_3\gg c_1,c_2$; $\bm{v}_{\rm des}$ is a desired velocity which has been calculated \textit{a priori} to produce a stable multi-hop trajectory; and ${\rm Bound}()$ is defined as
\begin{align*}
    {\rm Bound}(\alpha_k,\beta_k)=&[\max(\alpha_k-\overline{\alpha},\underline{\alpha}-\alpha_k,0) \\
    &+\max(\beta_k-\overline{\beta},\underline{\beta}-\beta_k,0)]
\end{align*}
for lower and upper bounds $\underline{\cdot},\overline{\cdot}$, respectively. The ${\rm Bound}()$ function penalizes the selection of leg impact angles outside the bounds of the training dataset range.

Due to the nonconvex nature of neural network optimization, Equation \eqref{eq: NN Optimization} may produce a local minimum solution which has high cost, resulting in deviation from the desired locomotion state. Occasionally, this deviation can constrain the capability of the robot to hop into a cell commanded from the IMDP. In this case, we optimize over a set of backup actions which are described in Section \ref{subsection: Low Level}.

The hopping robot IMDP abstraction along with the NN low-level optimization 
solve Subproblem \ref{subproblem:Abstract}.
\section{Synthesis Approach}\label{section:Synthesis}
In this section, we detail our approach to enforce LTL task satisfaction guarantees while simultaneously optimizing reward from optional tasks and learning system uncertainties.
\subsection{Multi-task PIMDP}
We first propose a Multi-task PIMDP structure which encodes the reward function, solving Subproblem \ref{subproblem:MT-PIMDP}.
\begin{definition}[Multi-task PIMDP]\label{def:PIMDP}
Let $\mathcal{I} = (Q, A, \check{T}, \hat{T},Q_0, O, L)$ be an IMDP abstraction of \eqref{eq: Flight Phase}, \eqref{eq: Impact Phase} and $\mathcal{A}=(S,s_0,O,\delta,F)$ be an DRA of LTL specifications. A Multi-task Product IMDP (MT-PIMDP) is defined as a tuple $\mathcal{P}^+ = \mathcal{I}\otimes\mathcal{A} =$\\ $(Q\times S, A, \check{T}', \hat{T}', Q\times s_0, F')$, where
\begin{itemize}
    \item $\check{T}': (q,s) \times A \times (q',s') :=
        \check{T}(q,\alpha,q') $ if $ s' \in \delta (s, L(q))$ and
        $0$ otherwise,
    \item $\hat{T}': (q,s) \times A \times (q',s') :=
        \hat{T}(q,\alpha,q') $ if $ s' \in \delta (s, L(q))$ and
        $0$ otherwise,
    \item $(q_0,\delta(s_0,L(q_0))) \in (Q\times S)$ is a set of initial states of $\mathcal{I}\otimes\mathcal{A}$,
    \item $F'=Q\times F=\{Q\times (G_1,B_1),\cdots,Q\times (G_n,B_n)\}$, where $G_i,B_i\subseteq S$ is the $i$th acceptance condition, and
    \item $W:Q\rightarrow\mathbb{R}$ is a reward function mapping PIMDP state-action pairs to their corresponding rewards.
\end{itemize}

The difference between the MT-PIMDP and the standard PIMDP introduced in Section \ref{section:Setup} is the addition of the reward function $W$. Figure \ref{fig:Overall Block Diagram} shows the interactions between the MT-PIMDP and the low-level components of the system.
\end{definition}
\subsection{Control Policy Synthesis}\label{subsection: Synthesis}
We now define the synthesis techniques which we will use for MT-PIMDP planning.
We first introduce the concept of control policies and adversaries.
\begin{definition}[Control Policy]
A control policy $\pi\in\Pi$ of a MT-PIMDP is a mapping $(Q\times S)^+\xrightarrow{}A$, where $(Q\times S)^+$ is the set of finite sequences of states of the MT-PIMDP.
\end{definition}
\begin{definition}[MT-PIMDP Adversary]
Given a MT-PIMDP state $(q,s)$ and action $\alpha$, an adversary $\zeta\in Z$ is an assignment of transition probabilities $T_{\zeta}'$ to all states $(q',s')$ such that
\begin{align*}
    \check{T}'((q,s),\alpha,(q',s'))&\leq T_{\zeta}'((q,s),\alpha,(q',s')) \\ &\leq\hat{T}'((q,s),\alpha,(q',s')).
\end{align*}

In particular, we use a \textit{minimizing} adversary, which realizes transition probabilities such that the probability of satisfying the specification is minimal, and a \textit{maximizing} adversary, which maximizes the probability of satisfaction.
\end{definition}

We denote the worst case probability of satisfaction under a maximizing control policy and minimizing adversary as
\begin{equation*}
     \check{P}_{\max}((q,s)\hspace{-2pt}\models\hspace{-2pt}\phi)=\max\limits_{\pi\in\Pi}\min\limits_{\zeta\in Z}P(w\hspace{-2pt}\models\hspace{-2pt}\phi \ |\ \pi,\zeta,w[0]=(q,s)),
\end{equation*}
and similarly the best case probability of satisfaction under a maximizing control policy and adversary as
\begin{equation*}
    \hat{P}_{\max}((q,s)\hspace{-2pt}\models\hspace{-2pt}\phi)=\max\limits_{\pi\in\Pi}\max\limits_{\zeta\in Z}P(w_i\hspace{-2pt}\models\hspace{-2pt}\phi \ |\ \pi,\zeta,w[0]=(q,s)).
\end{equation*}
We can now define failure states in the MT-PIMDP with respect to mandatory tasks as those which have probability $\hat{P}_{\max}=0$ of satisfying the LTL specification $\phi$.
\begin{equation*}
    \text{Failure States}
    =\{(q,s)\in Q\times S\ |\ \hat{P}_{\max}((q,s)\models\phi)=0\}
\end{equation*}
\begin{definition}[Nonviolating MT-PIMDP]\label{def:PIMDP Safety}
A MT-PIMDP $\mathcal{P}^+$ is \textit{nonviolating} with respect to an LTL specification $\phi$ if there exists no failure states in $\mathcal{P}^+$.
\end{definition}

We enforce a nonviolating control policy by restricting the allowable actions of the robot to a subset which stays within a nonviolating subgraph of the original MT-PIMDP. The algorithm for this operation has been detailed for the standard PIMDP case in \cite[Algorithm 1]{Jiang2023a}, and is applied similarly in this study.

In addition to maintaining a nonviolating control policy, we consider the problem of optimizing a given reward function. We apply a modified Q-learning algorithm which implements a state-ordering algorithm as in \cite{lahijanian_formal_2015} in order to resolve the IMDP transition probability intervals with respect to a minimizing adversary. This ensures that the computed reward values provide a lower bound on the true reward.

\subsection{Nonviolating Environment-Exploration Control Policy}\label{subsection: Sampling}
We now design a control policy for environment exploration.
To illustrate the full complexity of the problem, we consider two scenarios. In both cases, the LTL specifications $\phi$ are fully known, and there exist system uncertainties to be learned via exploration. In the first scenario, the reward function $W$ is fully known. Thus, the problem is to synthesize a controller which switches between a (known) reward-optimal action and an LTL-satisfying action to balance the satisfaction of both objectives. In the second scenario, the reward function $W$ is unknown. In this case, the switching policy must also explore to learn the reward function in addition to the LTL and reward-maximization objectives.

We first address the scenario with known reward. In this case, the initial control policy samples the environment to learn system uncertainties, simultaneously maximizing the reward gained. We first identify a nonviolating MT-PIMDP which can be safely explored. To this end, we will use \textit{maximal end components} \cite{baier_principles_2008}.

\begin{definition}[End Component \cite{baier_principles_2008}]
An \textit{end component} of a finite PIMDP $\mathcal{P}$ is a pair $(\mathcal{T},Act)$ with $\mathcal{T}\subseteq(Q\times S)$ and $Act:\mathcal{T}\rightarrow A$ such that
\begin{itemize}
    \item $\emptyset\neq Act(q,s)\subseteq A(q)$ for all states $(q,s)\in\mathcal{T}$,
    \item $(q,s)\in\mathcal{T}$ and $\alpha\in Act(q,s)$ implies $\{(q',s')\in\mathcal{T}\ |\ \hat{T}(q,\alpha,q'))>0,s'\in\delta(s,L(q))\}\subseteq\mathcal{T}$,
    \item The digraph $G_{(\mathcal{T},Act)}$ induced by $(\mathcal{T},Act)$ is strongly connected.
\end{itemize}
\end{definition}
\begin{definition}[Maximal End Component (MEC) \cite{baier_principles_2008}]
An end component $(\mathcal{T},Act)$ of a finite PIMDP $\mathcal{P}$ is \textit{maximal} if there is no end component $(\mathcal{T}^*,Act^*)$ such that $(\mathcal{T},Act)\neq (\mathcal{T}^*,Act^*)$ and $\mathcal{T}\subseteq \mathcal{T}^*$ and $Act(q,s)\subseteq Act^*(q,s)$ for all $(q,s)\in \mathcal{T}$.
\end{definition}

It is well known that PIMDP abstractions have the property that any infinite path will eventually stay in a single MEC. Thus, our goal is for the robot to reach an optimal MEC $(\mathcal{T}^*,Act^*)$ which has highest average reward on its states.
\begin{equation}\label{eq:Average Reward}
    (\mathcal{T}^*,Act^*)=\max_{(\mathcal{T},Act)}\frac{1}{|\mathcal{T}|}\sum_{(q,s)\in\mathcal{T}}W(q)
\end{equation}
where $|\mathcal{T}|$ is the cardinality of $\mathcal{T}$.

We design the following algorithm in order to select a MEC to cycle within. Given a set of candidate MECs $\{(\mathcal{T}_i,Act_i)\},i\in[1,n]$, we find the subset satisfying
\begin{equation*}
    \check{P}_{\max}((q_0,s_0)\hspace{-2pt}\models\hspace{-2pt}\eventually\mathcal{T}_i)=1,
\end{equation*}
\textit{i.e.} we can reach these MECs from the initial state of the system with worst case probability 1.
In case no MECs can be reached with probability 1, we immediately select the MEC with the highest $\check{P}_{\max}$. If multiple candidates remain, we calculate an optimal MEC $(\mathcal{T}^*,Act^*)$ as in Equation \eqref{eq:Average Reward}. 

We generate an LTL goal-reaching policy which reaches and stays in the optimal MECs $(\mathcal{T}^*,Act^*)$. We separately synthesize a reward-maximizing controller which selects the reward-maximizing action at each state, determined using offline Q-learning. 
At runtime, we switch between the goal-reaching and reward-maximizing policies. The probabilities $P_{LTL,ee}$ and $P_{RL,ee}$ of selecting the LTL and RL actions, respectively, under this environment exploration policy can be tuned to adjust the priority of the LTL and RL tasks.

Next, we consider the scenario in which the reward function for the optional tasks must be learned online. In this case, we create a reward-maximizing policy following a standard exploration-exploitation Q-learning sampling method. The LTL-satisfying policy reaches an optimal MEC as in the known reward case. We again select LTL and RL actions with probabilities $P_{LTL,ee}$ and $P_{RL,ee}$, respectively.

\subsection{Goal-Reaching Control Policy}\label{subsection: Goal Reaching}
Once the robot has learned the uncertainties sufficient to satisfy the LTL specifications $\phi$ with the desired probability $P_{sat}$, we need to synthesize a goal-reaching controller 
First, we perform a graph pruning operation to generate a \textit{satisfying} MT-PIMDP, leaving states $(q,s)$ which satisfy
\begin{equation*}
    \check{P}_{\max}((q,s)\hspace{-2pt}\models\hspace{-2pt}\phi)>P_{\rm sat}
\end{equation*} 
Given this satisfying MT-PIMDP, we synthesize an LTL goal-reaching policy using standard value-iteration techniques. For the known reward scenario, the robot synthesizes a reward-maximizing control policy using offline Q-learning. If the robot must learn the reward function, we use an exploration-exploitation approach analogous to the exploration policy of Section \ref{subsection: Sampling}. At runtime, we synthesize a unified goal-reaching policy by selecting LTL or RL actions based on the probabilities $P_{LTL,gr}$ and $P_{RL,gr}$, respectively. These follow a decaying exponential distribution
\begin{align}
    \label{eq: Epsilon Switching}P_{RL,gr}=Ce^{-\epsilon m}, \quad
     P_{LTL,gr}=1-P_{RL,gr}
\end{align}
where $C<1$ is a constant tuning the initial probability of selecting the RL policy, $\epsilon \in [0, 1]$ is a parameter which determines the rate of decay, and $m$ is the number of steps the robot has traversed under the goal-reaching policy. Thus, the robot initially takes more reward-maximizing steps and selects more LTL goal-reaching actions as the trajectory length increases. This policy devalues reward at later steps, similar to a discounted reward formulation. Together, the environment exploration and goal-reaching policies solve Subproblem \ref{subproblem:Planning}.

\begin{figure}[t]
    \centering
    \includegraphics[width=0.475\textwidth]{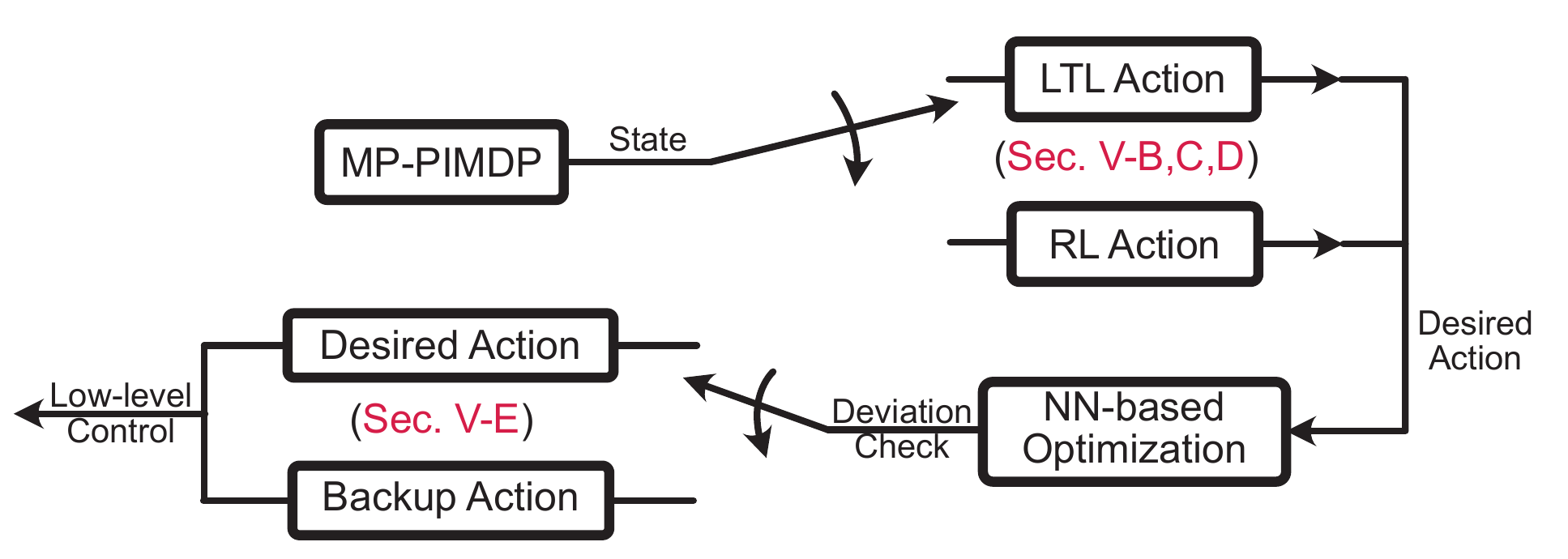}
    \caption{Diagram of the complete controller framework. A switch at the high level chooses between LTL goal-reaching and RL reward-maximizing actions, and a switch at the low level minimizes locomotion deviation by using a backup action if necessary.}
    \label{fig:Hopping Robot Switching}
    \vspace{-0.2in}
\end{figure}

\subsection{Low-level Execution}\label{subsection: Low Level}
As discussed in Section \ref{subsection: Learning Controller}, a desired MT-PIMDP action may have large cost when solving for the corresponding leg impact angles in Equation \eqref{eq: NN Optimization}, resulting in deviation from the targeted locomotion state. To address this issue, we propose a backup controller to provide a less deviating action. Recall that our synthesis algorithms in this section require an intermediate computation of a \textit{nonviolating} or \textit{satisfying} MT-PIMDP. 
If the desired action has a large error, we synthesize control inputs using the other available actions at that state in the MT-PIMDP and select the action with the smallest cost. 
The complete pipeline is shown in Figure \ref{fig:Hopping Robot Switching}.

\begin{figure*}[!t]
    \centering
    \includegraphics[width=0.99\textwidth]{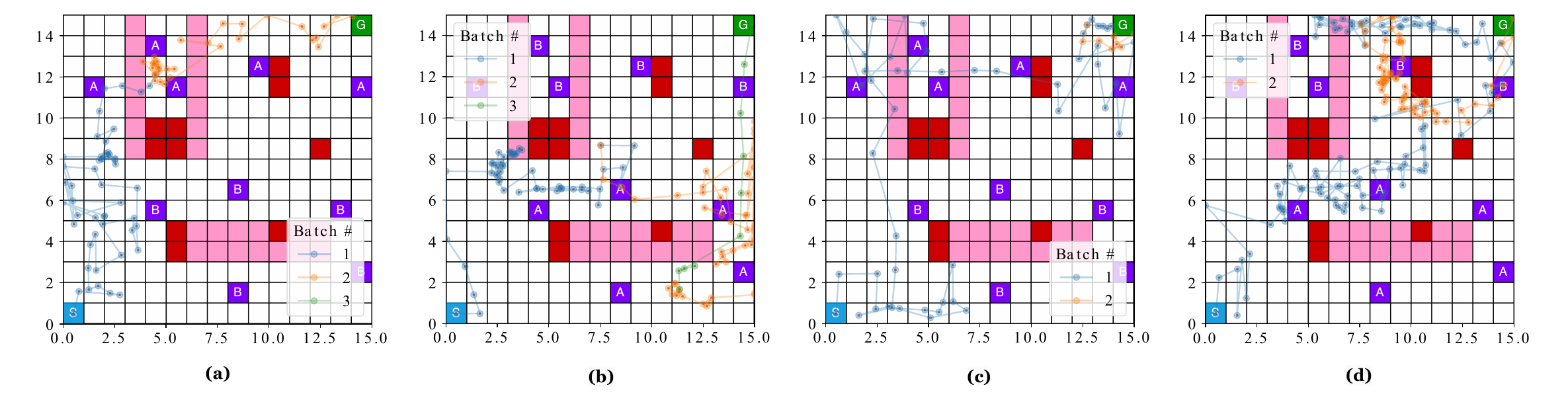}
    \caption{State space and results of the case studies. The initial region is in blue, the target region is green, and the hazard regions are red. Additionally, we have optional tasks with optional goals labeled in purple ("A" states with reward 20, and "B" states with reward 5), and weak hazards with penalty -0.5 labeled in pink. \textbf{(a)}: In this case, the reward function is known. \textbf{(b)}: A known reward environment similar to (a), but with the high-value "A" states now concentrated in the bottom right region of the environment. \textbf{(c)}: The robot traverses the same environment as in (a), but the reward function is unknown. The depicted trajectory shows the tenth run of the robot after it has learned the reward function over nine previous runs. \textbf{(d)}: The unknown reward version of (b), again depicting the tenth run through the environment.}
    \label{fig:CaseStudy1}
    \vspace{-0.2in}
\end{figure*}
\subsection{Analysis of Task Trade-off Behavior}

\begin{thm}[Trade-off Between Tasks]
Given the $\epsilon$-greedy switching policy \eqref{eq: Epsilon Switching} when applying the goal-reaching policy in Section \ref{subsection: Goal Reaching}, one has a trade-off between the speed of satisfying the LTL tasks and the maximization of the achieved reward.

\begin{proof}
    The main objective of this proof is to upper bound the number of steps $M_{LTL}$ required to satisfy an LTL specification $\phi$ as a function of the size of the state space and the value of the switching parameter $\epsilon$. To this end, we assume a worst case scenario in which the LTL and RL actions directly oppose each other. We also assume that all transitions in the MT-PIMDP are bidirectional: if there exists a transition from state $q_i$ to state $q_j$, there also exists a transition from $q_j$ to $q_i$. Additionally, we assume that all transitions reach their targeted state, \textit{i.e.}, there are no perturbations and the high-level action is perfectly tracked. 
    
    There exists a control policy to reach an accepting MEC which requires visiting each state in the MT-PIMDP at most once, as otherwise the policy is dependent on the trajectory history, violating the Markovian property of the MT-PIMDP. Thus, the maximum number of steps needed to reach the goal when executing only LTL actions is $|Q\times S|$, the size of the MT-PIMDP state space (\textit{i.e.}, the number of nodes of the product automaton). Then, we can ensure that the robot moves towards the goal in expectation when the probability $P_{LTL,gr}$ of selecting an LTL goal-reaching action exceeds the probability $P_{RL,gr}$ of selecting an RL reward-maximizing action. The number of steps $M_{\text{switch}}$ at which the condition $P_{LTL,gr}>P_{RL,gr}$ first holds is
    \begin{equation}
        Ce^{-\epsilon M_{\text{switch}}}=0.5 \implies M_{\text{switch}} = -\ln(0.5/C)/\epsilon
    \end{equation}
    When the threshold $M_{\text{switch}}$ is reached, we take more LTL than RL actions and move towards the goal. The expected movement (in number of states) is equal to the excess probability of selecting an LTL over an RL action
    \begin{equation}\label{eq: Movement}
       P_{LTL,gr}-P_{RL,gr}=1-P_{RL,gr}-P_{RL,gr}=1-2P_{RL,gr}.
    \end{equation}
    
    Finally, we can determine the upper bound $M_{LTL}$ on the number of steps needed to reach the goal by integrating Equation \eqref{eq: Movement} with lower bound $M_{\text{switch}}$ and evaluating when the expected steps towards the goal exceeds $|Q\times S|$. 
    \begin{align}\label{eq:Bound}
        &\int_{M_{\text{switch}}}^{M_{LTL}}(1-2P_{RL,gr})dm > |Q\times S|
    \end{align}
    Evaluating the left hand side,
    \begin{align}
       &\nonumber\int_{M_{\text{switch}}}^{M_{LTL}}(1-2Ce^{-\epsilon m})dm \\
       \nonumber &=(M_{LTL}-M_{\text{switch}})+\frac{2C}{\epsilon}(e^{-\epsilon M_{LTL}}-e^{-\epsilon M_{\text{switch}}}) \\ 
       \nonumber &=(M_{LTL}-M_{\text{switch}})+\frac{2C}{\epsilon}(e^{-\epsilon M_{LTL}}-e^{\epsilon\ln(0.5/C)/\epsilon}) \\
       &\label{eq:MLTL}=(M_{LTL}-M_{\text{switch}})+\frac{1}{\epsilon}\bigg[2Ce^{-\epsilon M_{LTL}}-1\bigg]
    \end{align}
    For $M_{LTL}>M_{\text{switch}}$, we see that the second term of \eqref{eq:MLTL} is always negative. Increasing $\epsilon$ decreases the magnitude of the term since the $1/\epsilon$ term dominates $e^{-\epsilon M_{LTL}}$. 
    Thus, increasing $\epsilon$ reduces the magnitude of $M_{LTL}$ needed to satisfy the bound \eqref{eq:Bound}. We have then proved that increasing $\epsilon$ results in faster expected LTL task satisfaction. In the general case of non-ideal MT-PIMDP transitions, we argue that a similar trend holds since the dependence is on the proportion of LTL and RL actions taken.
    
    The RL reward-maximizing action is selected with probability $Ce^{-\epsilon m}$. Thus, the cumulative proportion of RL actions selected is
    \begin{equation}
        \frac{1}{M_{LTL}}\int_0^{M_{LTL}} Ce^{-\epsilon m}dm = \frac{C}{M_{LTL} \epsilon}(1-e^{-\epsilon M_{LTL}})
    \end{equation}
    This is a monotonically decreasing function in $\epsilon$, matching our intuition that increasing $\epsilon$ 
    will decrease the number of RL actions taken and therefore the expected reward.
\end{proof}
\end{thm}

\section{Case Study}
Consider the hopping robot modeled as in Equations \eqref{eq: Flight Phase}-\eqref{eq: Impact Phase} in an environment of size $15\times15$ meters. This environment is partitioned into hyperrectangular regions with length 1 meter.
Within the state space, we have one goal region with the atomic proposition \texttt{Goal} and a set of hazard regions labeled with \texttt{Haz}. These yield the LTL specification
\begin{equation}\label{eq:LTL}
    \phi_1 = \lnot \texttt{Haz} \ \until \ \texttt{Goal}.
\end{equation}
Additionally, we have two classes of optional tasks $A$ and $B$, where $A$ regions have a reward of 20 and $B$ regions have a reward of 5. There also exist weak hazard regions which induce a small traversal penalty. There is no reward or penalty associated with the LTL hazard and goal regions, so the LTL and RL tasks are independent of each other. Figure \ref{fig:CaseStudy1} illustrates the environmental setup.

The system uncertainty is caused by the error inherent in the neural network controller of Section \ref{subsection: Learning Controller}, as well as additional state-dependent motion perturbation and stochastic noise injected into the system. The artificial state-dependent perturbation is drawn independently in the $x$ and $y$ dimensions from Gaussian processes. The stochastic noise $\nu$ is independently drawn from two truncated Gaussian distributions for the same dimensions, and both with $\sigma_\nu = 0.1$ and bounded support $[-0.2,0.2]$.
We estimate the unknown dynamics with two sparse Gaussian processes 
and set the number of inducing points $\eta=50$. 
In all cases, we set $P_{\rm sat}=0.65$ as the desired probability for LTL satisfaction.

All case studies were run on a 4.5 GHz AMD Ryzen 7700X with 64 GB of RAM and a Nvidia RTX 3090 GPU.
\subsection{Known Reward Case Studies}
We first present two scenarios of the hopping robot navigating through environments in which the reward function is \textit{a priori} known. The first scenario, depicted in Figure \ref{fig:CaseStudy1}(a), concentrates "A" states in the upper-left region of the environment, and "B" states in the lower-left region. There is a penalty of -0.5 for traversing pink states. The robot explores towards the region with high-reward "A" states, continuing on towards the LTL goal while avoiding hazard states. To illustrate the effect of the reward function on the robot's control policy, we test a second environment in which the positions of the "A" and "B" states are reversed, as shown in Figure \ref{fig:CaseStudy1}(b). As expected, the robot changes its trajectory in order to target the new high-reward states while still satisfying the LTL specification. In both experiments, the robot traverses the environment in "batches" of 50 steps and collects trajectory data. After each batch, the robot retrains the GPs and updates its control policy. This cycle continues until the robot reaches the goal. 

\subsection{Unknown Reward Case Studies}
We now present experiments in the same environments, but where the reward function is \textit{a priori} unknown. In this case, more exploration of the environment is necessary in order to learn the reward simultaneously with the motion perturbations. In order to demonstrate the full efficacy of the Q-learning procedure, we perform multiple complete (\textit{i.e.}, until the LTL goal region is reached) runs of the robot through the environment, allowing the robot to retain knowledge of the learned reward function between runs but resetting its knowledge of the dynamical uncertainties. Thus, the robot improves its estimation of the reward function through the runs while still being forced to explore in order to relearn motion perturbations. 

Figure \ref{fig:CaseStudy1}(c) shows the tenth run of the robot learning the environment depicted in Figure \ref{fig:CaseStudy1}(a). The robot follows a similar trajectory as in the known reward case, moving towards the high-reward states in the upper-left region and then on towards the goal. Figure \ref{fig:CaseStudy1}(d) shows the unknown reward version of Figure \ref{fig:CaseStudy1}(b). In this case, the robot has learned a more inaccurate version of the true reward function and thus traverses a mix of high- and low-reward states on its way to the goal.

\subsection{Effect of Switching Parameters}
Each entry in Table \ref{table:Switching} shows results of the reward achieved (left value) and number of steps (right value) averaged across ten complete runs through the scenario of Figure \ref{fig:CaseStudy1}. The entries in the table have varying values of the switching parameter $\epsilon$, the fixed probability $P_{RL,ee}$ of selecting an RL action during the environment exploration policy, and the initial probability $P_{RL,gr}$ (i.e., $C$ when $m =0$) of selecting an RL action in the goal-reaching policy. For the sake of simplicity in presenting our results, we set $P_{RL,ee}=C$ for all cases, since from Equation \eqref{eq: Epsilon Switching} this enforces $P_{RL,ee}=P_{RL,gr}$ at the initial step $m=0$. In practice, this constraint is unnecessary, and a general setup can be considered.

The results of Table \ref{table:Switching} indicate that $P_{RL,ee}$ and $C$ have a large effect on the achieved reward, as higher probabilities of selecting RL actions generate significantly higher rewards. However, these parameters have no clear correlation with the trajectory length of the runs. On the other hand, $\epsilon$ has a large correlation with trajectory length, as a faster decay decreases the number of steps required to reach the LTL goal for $P_{RL,ee},C=\{0.25,0.50\}$, with the effect becoming more pronounced as $P_{RL,ee}$ and $C$ decrease. This matches our intuition that decreasing the proportion of RL actions selected more quickly should allow the robot to reach the goal in fewer steps, since the robot selects more LTL actions given the same length of trajectory. Additionally, $\epsilon$ shows some weak correlation with the achieved reward, with a faster decay resulting in a lower reward. Thus, tuning the combined parameter set $\{P_{RL,ee}, C, \epsilon\}$ allows for significant control over the behavior of the unified reward-seeking and goal-reaching control policy.
\begin{table}[t]
\centering
\caption{Comparison of Rewards Achieved (Left)/Trajectory Length (Right) as a Function of Switching Parameters}
\begin{tabular}{|c|c|c|c|}\hline
\diaghead{\theadfont Diag Columnmabc}
{$P_{RL,ee}$, $C$}{\\ $\epsilon$}&
\thead{$0.0025$}&\thead{$0.0050$}&\thead{$0.0075$}\\    \hline
$0.25$ & 61.2/98.0 & 64.4/80.6 & 54.6/\textbf{65.1}\\    \hline 
$0.50$ & 112.9/87.2  & 88.9/72.8 & 84.7/71.2\\    \hline
$0.75$ & 136.4/84.7 & \textbf{139.75}/83  & 129.6/85.3\\    \hline 
\end{tabular}
\vspace{-0.2in}
\label{table:Switching}
\end{table}
\subsection{Simulation Implementation}
In the video\footnote{\href{https://www.youtube.com/watch?v=USTBhrJmePM}{https://www.youtube.com/watch?v=USTBhrJmePM}} accompanying this article, we showcase 3D visualizations of example runs of the hopping robot navigating through the environment of our case studies. These visualizations are built in part on the Xpp visualization library \cite{xpp_ros} and are displayed in an RViz environment. The leg actuation of the visualized model (with a prismatic DOF on the leg) slightly differs from the model considered in this work, but our model remains valid as the SLIP model used for planning considers only CoM and foot positions.

\section{Conclusion}
In this work, we designed planning algorithms which allow a hopping robot to simultaneously satisfy mandatory goal-reaching tasks specified using LTL and optimize optional exploration tasks specified using reward functions in the presence of system uncertainties. These results can be extended to other robotic systems which can be abstracted as an IMDP, such as bipedal walking and wheeled robots with system and environmental uncertainties. 
Future work will apply these methods to our bipedal robot Digit, exploring the broad class of real-world tasks enabled using the proposed framework.

\bibliographystyle{IEEEtran}
\bibliography{references,ref}

\end{document}